\documentclass[10pt,twocolumn,letterpaper]{article}

\usepackage{iccv}
\usepackage{times}
\usepackage{epsfig}
\usepackage{graphicx}
\usepackage{amsmath}
\usepackage{amssymb}
\usepackage{booktabs}
\usepackage{multicol}
\usepackage{multirow}
\usepackage{wrapfig}
\usepackage{enumitem}
\usepackage{array}
\usepackage{gensymb}
\usepackage{siunitx}
\usepackage{setspace}

\newcolumntype{?}{!{\vrule width 1pt}}

\usepackage[breaklinks=true,bookmarks=false,colorlinks=true,backref=page]{hyperref}

\iccvfinalcopy % *** Uncomment this line for the final submission

\def\iccvPaperID{4383} % *** Enter the ICCV Paper ID here

\ificcvfinal\fi
\newcommand{\mypara}{\noindent\textbf}

\begin{document}

%%%%%%%%% TITLE
\title{Act the Part:\\Learning Interaction Strategies for Articulated Object Part Discovery}

\author{Samir Yitzhak Gadre$^{1}$\hspace{10mm}Kiana Ehsani$^2$\hspace{10mm}Shuran Song$^1$\\
$^1$ Columbia University\hspace{10mm}$^2$ Allen Institute for AI\\
\href{https://atp.cs.columbia.edu}{https://atp.cs.columbia.edu}
}

\maketitle
% Remove page # from the first page of camera-ready.
\ificcvfinal\thispagestyle{empty}\fi

%%%%%%%%% ABSTRACT
\begin{abstract}

People often use physical intuition when manipulating articulated objects, irrespective of object semantics.
Motivated by this observation, we identify an important embodied task where an agent must play with objects to recover their parts.
To this end, we introduce Act the Part (AtP) to learn how to interact with articulated objects to discover and segment their pieces.
By coupling action selection and motion segmentation, AtP is able to isolate structures to make perceptual part recovery possible without semantic labels.
Our experiments show AtP learns efficient strategies for part discovery, can generalize to unseen categories, and is capable of conditional reasoning for the task.
Although trained in simulation, we show convincing transfer to real world data with no fine-tuning.
A summery video, interactive demo, and code will be available at \href{https://atp.cs.columbia.edu}{https://atp.cs.columbia.edu}.
\end{abstract}

%%%%%%%%% BODY TEXT
\section{Introduction}

How do people and animals make sense of the physical world?
Studies from cognitive science indicate observing the consequences of one's actions plays a crucial role \cite{held_1963, roberson_2001, renee_2004}.
Gibson's influential work on affordances argues visual objects ground action possibilities \cite{gibson_1979}.
Work from Tucker \textit{et al.} goes further, suggesting what one sees affects what one does \cite{tucker_1998}.
These findings establish a plausible biological link between seeing and doing. 
However, in an age of data-driven computer vision, static image and video datasets \cite{russakovsky_2015, lin_2015, abuelhaija_2016} have taken center stage.

\begin{figure}[t]
    \centering
    \includegraphics[width=\linewidth]{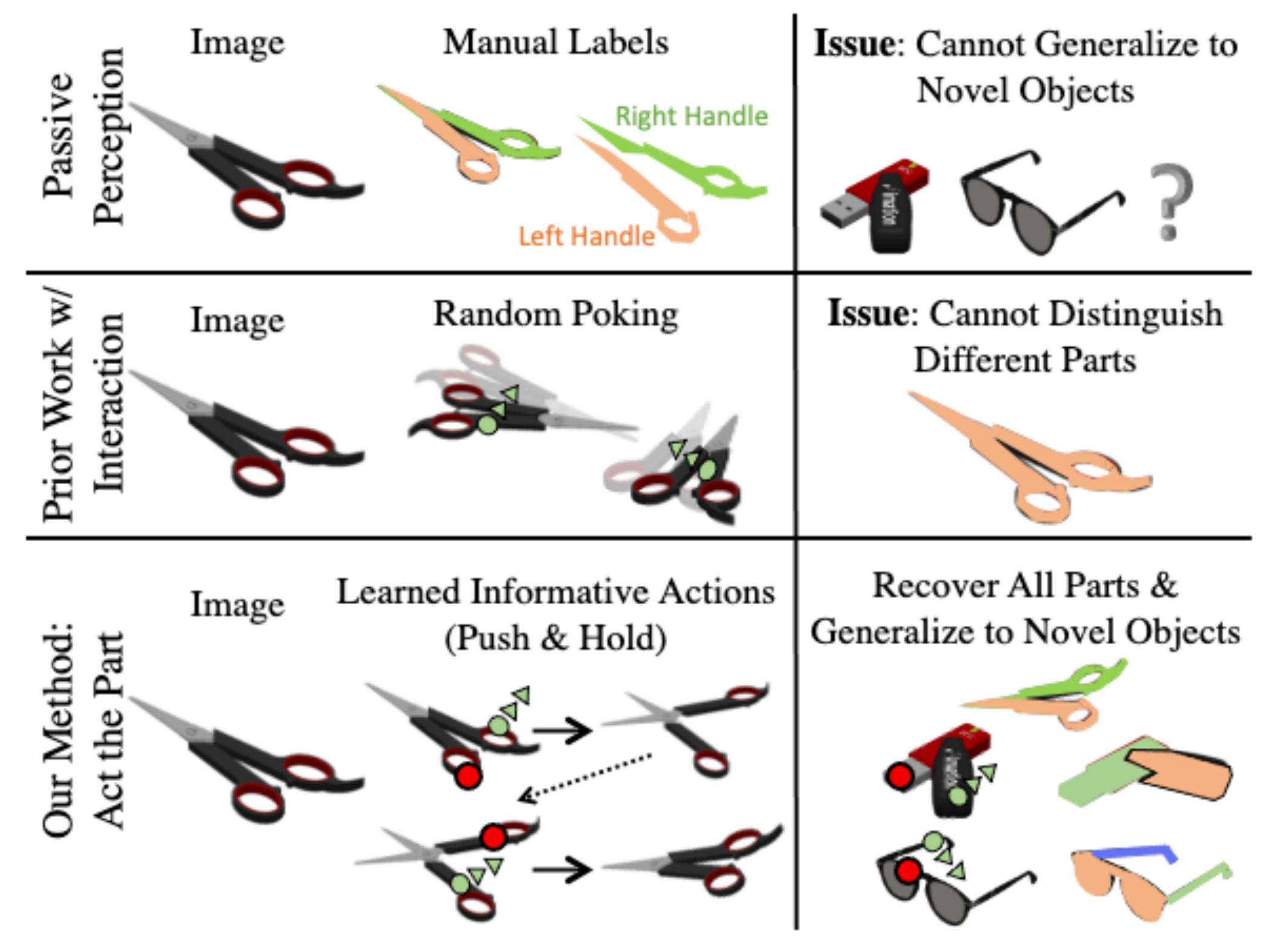} 
    \caption
    {\textbf{Interaction for Part Discovery.}
    Passive part segmentation algorithms require detailed annotation and cannot generalize to new categories. 
    While motion can help discover new objects, prior work cannot infer actions for understanding individual parts.
    Our work, Act the Part, learns interaction strategies that expose parts and generalize to unseen categories.}
    \label{fig:teaser}
\end{figure}

In this paper, we aim to elucidate connections between perception and interaction by investigating articulated object part discovery and segmentation.
In this task, an agent must recover part masks by choosing strategic interactions over a few timesteps.
We do not assume dense part labels or known kinematic structure \cite{abbatematteo_2020, li2020categorylevel}.
We also do not interact randomly \cite{Pathak_2018}.
Rather, we \textit{learn} an agent capable of \textit{holding} and \textit{pushing}, allowing us to relax the assumption that objects are fixed to a ground plane \cite{mo_2021}.
Our task and approach novelty are highlighted in Fig. \ref{fig:teaser}.

Segmentation from strong supervision and random interaction is widely studied; however, creating \textit{informative} motion to enable category level generalization while relaxing supervision is less explored in the community.
We identify the following hurdles, which make this direction salient and difficult.
Motion cannot be assumed in a scene as objects seldom move spontaneously.
Even with agent interaction, not all actions create perceivable motion to give insight about articulation.
Actions might activate only a small number of parts, so diversity of action and aggregation of potentially noisy perceptual discoveries is necessary.
Generalization of interaction and perception to unseen categories without retraining or fine-tuning is also desirable.
These facets are often overlooked in prior work but are at the heart of this paper.

To address these challenges, we introduce \textbf{Act the Part (AtP)}, which takes visual observations, interacts intelligently, and outputs part masks.
Our key insight is to couple action selection and segmentation inference.
Given an RGB input image and the part segmentation belief, our interaction network reasons about where to hold and push to move undiscovered parts. 
By reasoning about changes in visual observations, our perception algorithm is able to discover new parts, keep track of existing ones, and update the part segmentation belief.  

We evaluate our approach on eight object categories from the PartNet-Mobility dataset \cite{chang_2015, Mo_2019, Xiang_2020} and a ninth multilink category, which we configure with three links.
Our experiments suggest:
(1) AtP learns effective interaction strategies to isolate part motion, which makes articulated object part discovery and segmentation possible.
(2) Our method generalizes to unseen object instances and categories with \textit{different} numbers of parts and joints.
(3) Our model is capable of interpretable conditional reasoning for the task---inferring where and how to push given arbitrary hold locations.

We also demonstrate transfer to real images of unseen categories (without fine-tuning) and introduce a toolkit to make PartNet-Mobility more suitable for future research.
\section{Related Work}
Our approach builds on existing work in interactive perception \cite{bohg_2017}, where visual tasks are solved using agent intervention.
We also position our work alongside existing methods in articulated object understanding.

\mypara{Interactive Perception for Rigid Objects.}
Instance segmentation of rigid objects from interaction is well studied \cite{Fitzpatrick_2003,bjorkman_2010, van_hoof_2014, Pajarinen_2015, Byravan_2017, Pathak_2018, Eitel_2019}.
Similar work infers physical properties \cite{pinto_2016, xu_2019} and scene dynamics \cite{nematoli_2020, xu_2020, tung_2020}.
These approaches typically employ heuristic or random actions.
In contrast, we \textit{learn} to act to expose articulation.

For learning interaction strategies, Lohmann \textit{et al.} \cite{Lohmann_2020} learn to interact with rigid objects to estimate their segmentation masks and physical properties.
Yang \textit{et al.} \cite{Yang_2019} learn to navigate to recover amodal masks. 
These algorithms do not change object internal states in structured ways for articulated object part discovery. 

There is also work that leverages multimodal tactile and force inputs \cite{Chu_2015}.
Inspired by this work, we explore using touch feedback in our learning loop.
However, we assume only binary signals (e.g., the presence of shear force), which is easier to obtain in real world settings.

\mypara{Passive Perception for Object Structures.}
Existing work extracts parts from pairs of images \cite{Yan_2006, xu2019Unsupervised}, point clouds \cite{Yi_2019} or videos \cite{Schmidt_2014, martin_2014, liu2020geometric}. 
In these settings, agents do not have control over camera or scene motion. 
While the assumption that structures move spontaneously is valid for robot arms or human limbs, the premise breaks down when considering inanimate objects.
Even when motion exists, it is not guaranteed to give insight about articulation.
We address these issues by learning how to create informative motion to find and extract parts.

Other work tackles part segmentation from a single image  \cite{Wang_2015, TsogkasKPV15, hung2019scops, Lee2020CameratoRobotPE, abbatematteo_2020, li2020categorylevel} or point clouds \cite{qi2017pointnet, qi2017pointnet2, Yue_2019, huang_2021}.
These algorithms are trained with full supervision (e.g., pixel labels) or assume strong category-level priors (e.g. known kinematics or single networks per category).
In contrast, our approach uses flow and touch feedback as supervision and makes no class specific assumptions.
As a result, we are able to learn \textit{a single model} for \textit{all} our object categories, which encompass diverse kinematic structures.

\mypara{Interactive Perception for Articulated Objects.}
In traditional pipelines, agents are carefully programmed to execute informative actions to facilitate visual feature tracking \cite{Sturm_2011, Katz_2013, Pillai_2014}.
Other classical approaches improve on action selection for downstream perception \cite{Barragan2014InteractiveBI, otte_2014, Hausman2015ActiveAM}.
However, these methods assume known object structure, which is used to design heuristics.
In contrast, we employ an end-to-end learnable framework, which allows learning actions directly from pixels without known object models.

Recently, Mo \textit{et al.} \cite{mo_2021} present a learnable framework to estimate action affordences on articulated objects from a single RGB image or point cloud.
However, they do not consider using their learned interactions for multistep part discovery and segmentation.  

\section{Approach}

Our goal is to learn how to interact with articulated objects to discover and segment parts without semantic supervision.
This poses many technical challenges:
(1) With repetitive actions, an agent may not explore all parts.
(2) Actions resulting in rigid transformations are undesirable.
(3) Erroneous segmentation makes tracking parts over time difficult.
To begin exploring these complexities, we consider articulated objects in table-top environments.

\begin{figure}[t]
\includegraphics[width=\linewidth]{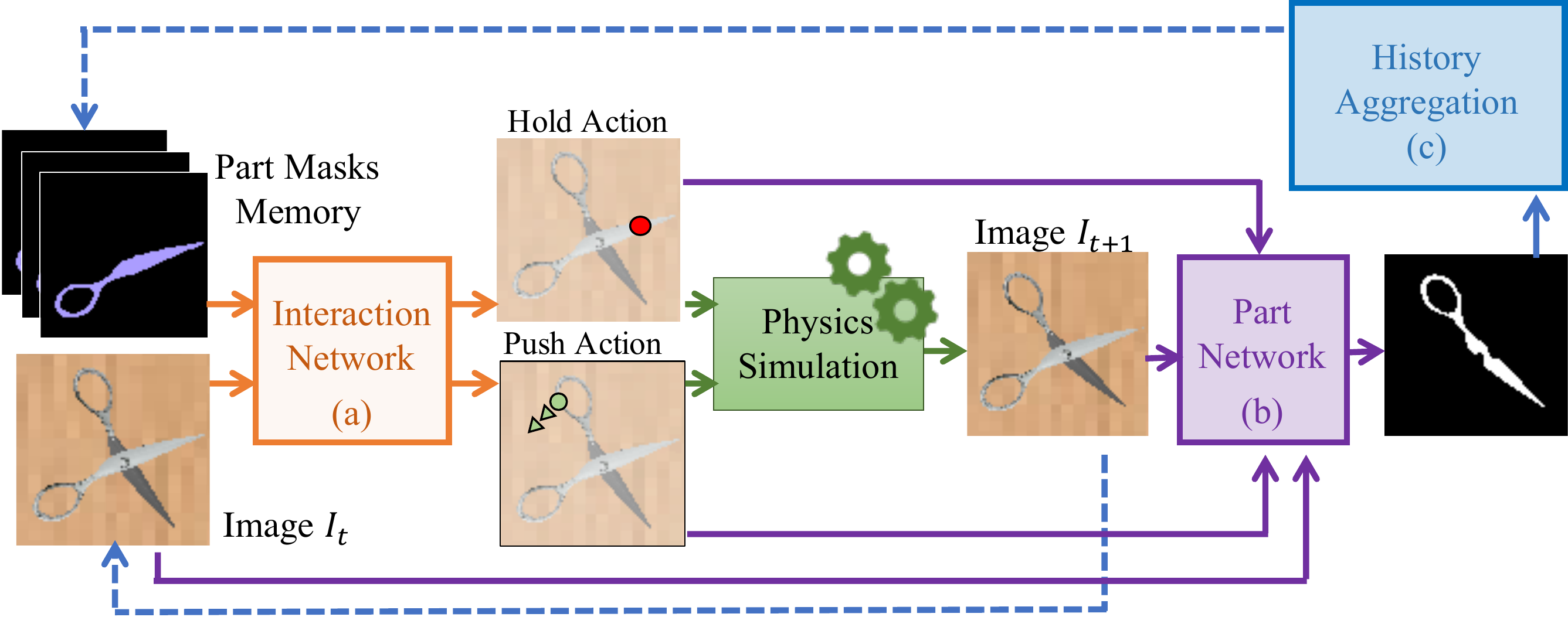} 
    \caption {\textbf{Model overview.}
    \textbf{(a)} The interaction network computes hold and push from an image observation and current part memory.
    The physics simulator gives the action effects yielding the next observation.
    \textbf{(b)} The part network takes the action and image observations to infer the motion masks for the part that moved, one aligned to the current frame and one to the next frame. 
    \textbf{(c)} The history aggregation module incorporates newly discovered parts and updates existing parts in the memory.}
    \label{fig:overall}
\end{figure}

First, we formally define the task and environment details (Sec. \ref{sec:task}).
We then explain the three components of our approach: an interaction network (Sec. \ref{sec:interaction}) to determine what actions to take, a part network (Sec. \ref{sec:part}) to recover masks from motion, and a history aggregation algorithm (Sec. \ref{sec:history}) to keep track of discovered parts.
Finally, we explain the reward formulation (Sec. \ref{sec:reward}) and combine our modules to present the full pipeline (Sec. \ref{sec:full_pipeline}), Act the Part (AtP).
Our approach is summarized in Fig. \ref{fig:overall}.

\subsection{Problem Formulation}
\label{sec:task}
\mypara{General Setting.}
Let $\mathcal{O}$ denote a set of articulated objects, each with $n \leq N$ parts.
At each timestep $t$, an agent gets an observation $I_t \in \mathbb{R}^{H \times W \times C}$, and executes an action $a_t \in \mathcal{A}$ on an object $o \in \mathcal{O}$, where $\mathcal{A}$ is the set of all possible actions.
Additional sensor readings $s_t \in \mathbb{R}^l$ complement visual perception.
The action results in the next observation $I_{t+1}$.
Given the sequence of $T$ observations, sensor readings, and actions, the goal is to infer part mask $\mathcal{M}_T \in \{1, 2, ..., N+1\}^{H \times W}$, where each pixel is assigned a value corresponding to $N$ part labels or background. 

\mypara{Task Details.}
We consider $\mathcal{O}$ to be a set of common household objects with $n \leq 3$ parts, $T=5$, $W,H=90$, and $C=3$ (RGB).
All objects have revolute joints and no fixed base link.
Each $a \in \mathcal{A}$ represents a tuple: an image pixel to hold, another pixel to push, and one of eight push directions.
The directions are discretized every 45$\degree$ and are parallel to the ground plane.
We take $s_t \in \{0, 1\}^3$, representing binary signals for detecting contact on the hold and push grippers and a binary sheer force reading on the hold gripper to emulate touch.

\mypara{Environment Details.}
To enable large-scale training and ground truth part segmentation (for benchmarking only), we use a simulated environment.
However, we also show our model generalizes to real-world images without fine-tuning.
Our simulation environment is built using PyBullet \cite{coumans2016pybullet}  with Partnet-Mobility \cite{chang_2015, Mo_2019, Xiang_2020} style dataset assets. 

Our environment supports two generic actions.
First, a hold action parameterized by its location and implemented as a fixed point constraint between the gripper and a part.
Second, a push action parameterized by the location and the direction of the applied force.
Actions are easily extensible to facilitate future 2D and 3D object interaction research.

\subsection{Learning to Act to Discover Parts}
\label{sec:interaction}
\begin{figure}[t]
    \begin{center}
    \includegraphics[width=\linewidth]{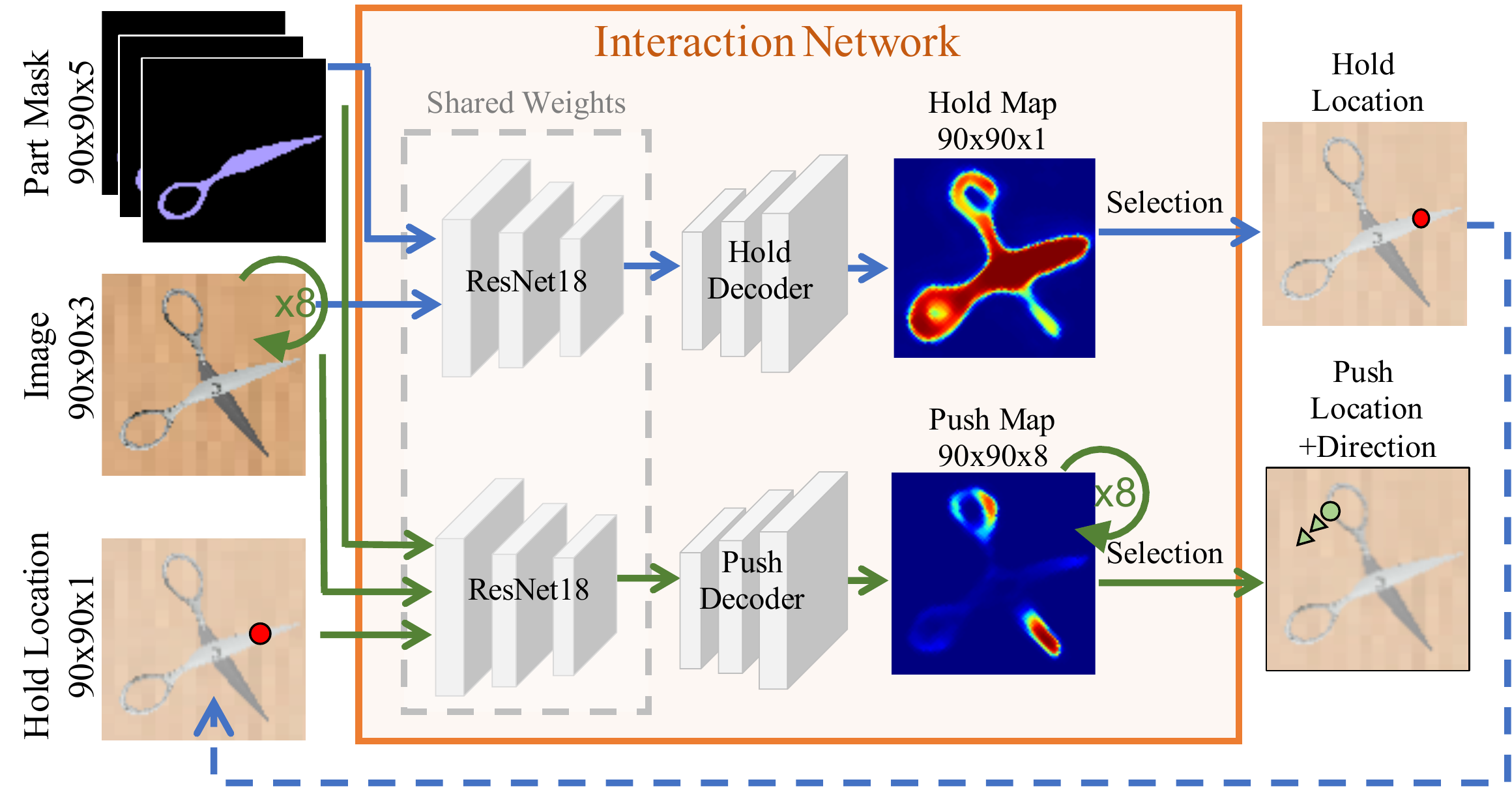}
    \end{center}
    \caption
    {\textbf{Interaction network.}
    Given an image and the current belief of part segmentation, our network predicts a hold and a push conditioned on the hold.}
    \label{fig:policy}
\end{figure}

Given a visual observation of an object, we want to create motion by interacting to expose articulation.
We give the agent two sub-actions every timestep: hold and push.
The action space directly motivates network and reward design.

\mypara{Conditional Bimanual Action Inference.}
The interaction task reduces to finding pixels to hold and push and determining the push direction.
To decrease the search space, we discretize the push direction into eight options (45$\degree$ apart).
We consider a constant magnitude push force parallel to the ground plane.
We condition the predicted push location and direction on the predicted hold location.
This allows us to synchronize sub-actions without considering all pairs.

\mypara{Interaction Network.} 
At every timestep, we predict one step pixel-wise reward for holding and pushing at the spatial resolution of the input image, similar to Zeng \textit{et al.} \cite{zeng2018learning}.
As shown in Fig. \ref{fig:policy}, we use a shared ResNet18 \cite{He2016} with two residual decoder heads wired with U-Net \cite{Ronneberger2015} skip connections.
At each timestep $t$, we have a current belief about the part segmentation.
This is represented as a part memory $V_t \in \{0, 1\}^{H \times W \times N}$, where each channel encodes a different part mask.
Given an image $I_t$ and $V_t$, the network predicts a hold reward map $H_t \in [0, 1]^{H \times W}$, where each entry estimates the reward for holding that pixel.
We uniformly sample one of the top $k=100$ pixels from $H_t$ as the hold location.
Sampling encourages optimization over the top $k$ actions, which we notice is necessary for the model to learn effective strategies.

Since we wish to infer pushing based on holding, we encode the hold as a 2D Gaussian $h_t$ centered at the hold location with standard deviation of one pixel \cite{jakab2018unsupervised}.
In doing so, we can pass the hold location in a manner that preserves its spatial relationship to $I_t$ and $V_t$.
To predict the push reward maps, we pass eight rotations of $I_t$, $V_t$, and $h_t$---every $45\degree$---through the push network.
The rotations allow the network to reason implicitly about pushing in all eight directions, while reasoning explicitly only about pushing right \cite{zeng2018learning}. 
We consider the output map with the largest reward, whose index encodes the push direction, and sample uniformly from the top $k=100$ actions to choose the push location.
An emergent property of our network is conditional reasoning, where hold locations can come from anywhere and the network still reasons about a synchronized push.
We demonstrate this capability on real world data in our experiments (Sec. \ref{sec:real}).

During training, we rollout the current interaction network for seven timesteps for each training instance.
The data and the corresponding reward (Sec. \ref{sec:reward}) for the last 10 iterations of rollouts are saved in a training buffer.
We use pixel-wise binary cross entropy loss to supervise the hold and push reward maps.
All pairs of frames, executed actions, and optical flow ground truth are saved to disk to train the part network described next.

\subsection{Learning to Discover Parts from Action}
\label{sec:part}

\begin{figure}[t]
\begin{center}
\includegraphics[width=0.97\linewidth]{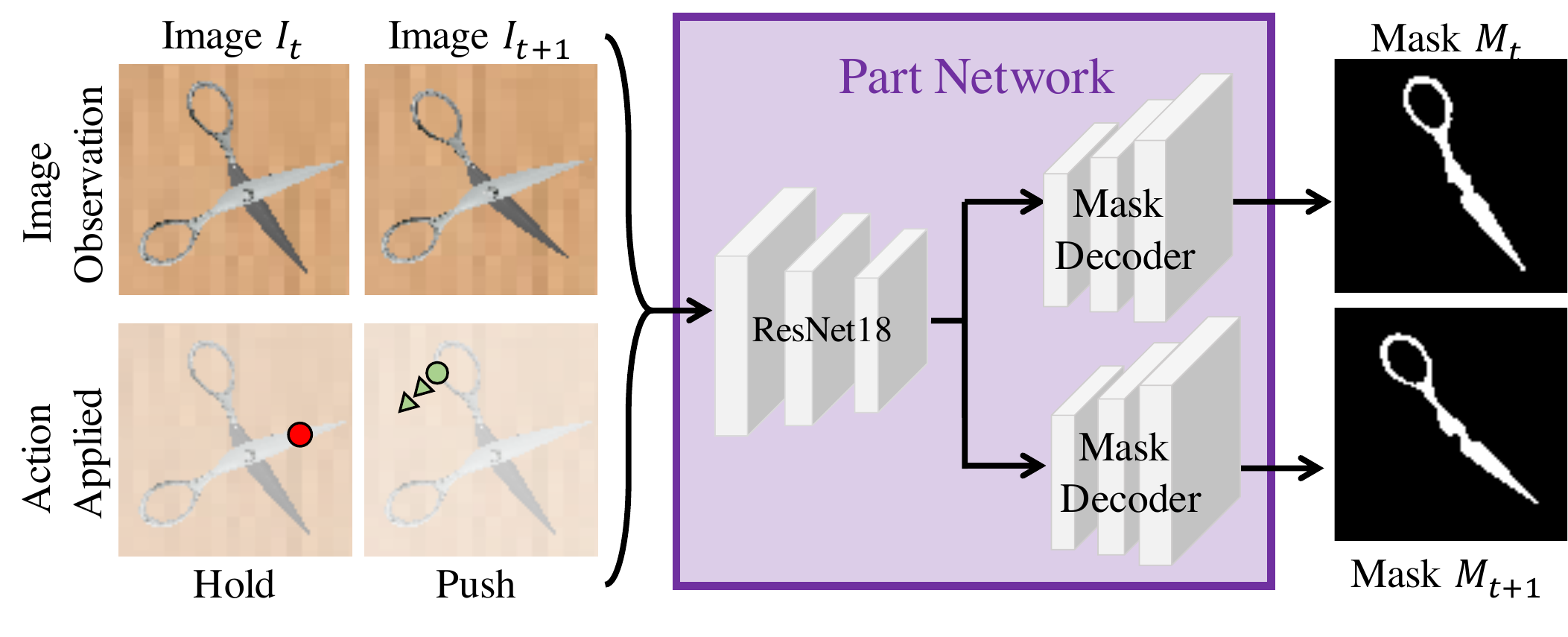}
\end{center}
\caption
{
\textbf{Part network.} Given a pair of observations and the action that caused the change, this network predicts motion masks aligned to each observation.
}
\label{fig:structure}
\end{figure}

After an action is executed, we wish to recover the moved part.
To do so, we create a part network to predict two masks for the pixels that moved---one mask $M_t$ aligned to the current frame and the other $M_{t+1}$ to the next frame.

\mypara{Part Network.}
Our part network (Fig. \ref{fig:structure}) takes the observations before and after the interaction.
Additionally we pass in the hold location $h_t$ and a spatial encoding $p_t$ of the push location and direction.
$p_t$ has a 2D Gaussian centered at the push pixel, analogous to $h_t$. To encode direction, we add Gaussians of smaller mean value in the direction of the push, forming a trail.
The network is comprised of a shared encoder with two decoder heads to predict $M_t$ and $M_{t+1}$.
Using consistent forward and backward flow collected during interaction network training, we threshold at zero to acquire target motion masks.
We supervise predictions using binary cross-entropy loss.

\subsection{History Aggregation}
\label{sec:history}
We introduce a history aggregation algorithm to updated part memory $V$, based on predicted $M_t$ and $M_{t+1}$.
Our algorithm classifies the type of update into four categories: (1) no movement, (2) finding a new part, (3) moving an existing part, (4) entangling parts.
These labels are used to decide how to update $V$ and influence the reward (Sec. \ref{sec:reward}).

\mypara{New Part.}
If $M_t$ does not overlap significantly with any channels in $V$, it is likely to be a new part.
A free channel $c$ is assigned: $V^c \gets M_{t+1}$.
If there is significant overlap between $M_t$ and a mask $V^c$, relative only to the area of $M_t$, there is indication two parts are assigned to $V^c$ that must be split: $V^c \gets V^c - (V^c \cap M_t)$ and $V^{c+1} \gets M_{t+1}$.
Finding a new part is the most desirable case.

\mypara{Existing Part.}
If there is significant overlap between $M_t$ and a mask $V^c$, relative to the areas of both $M_t$ and $V^c$, we execute the update: $V^c \gets M_{t+1}$.
This case is less desirable than discovering a new part.

\mypara{Entangled Parts.}
If there is significant overlap between $M_t$ and a mask $V^c$, relative to the area of only $V^c$, it suggests our action is entangling movement of more than one part.
During training; $V^c \gets M_{t+1}$.
During testing, we use Iterative Closest Point (ICP) to get the correspondences between $V^c$ and $M_{t+1}$, yielding $T \in SE(2)$, to execute the updates: $V^c \gets (T \circ V^c) \cap M_{t+1}$, then $V^{c+1} \gets M_{t+1} - V^c$.
Entangled part actions are the least desirable, as reflected in our reward described next.

For more details on handling edge cases (e.g., all channels being filled at allocation time), refer to Appx. \ref{appx:history_aggregation}.
\subsection{Reward}
\label{sec:reward}

During training, reward for the interaction network is determined from the optical flow, touch feedback, and history aggregation case. The reward conditions and values are shown in Tab. \ref{tab:reward}.

\begin{table}
 \scriptsize  
  \centering
  \tabcolsep=0.1cm
  \begin{tabular}{ccccc}         
  \toprule
         Optical Flow & Touch Sensor & Part Memory & Hold Reward & Push Reward\\
    \midrule
    x & 1/0 & - & N/A & 0\\
    \checkmark & 1 & New part & 1 & 1 \\
    \checkmark & 1 & Existing part & .5 & .5 \\
    \checkmark & 0 & - & 0 & N/A \\
    \checkmark & 1 & Entangled part & 0 & N/A \\
    \bottomrule
  \end{tabular}
  \caption{ \textbf{Reward Calculation.} N/A indicates no backpropagation due to insufficient information. For more details refer to Appx. \ref{appx:reward}.}
  \label{tab:reward}
\end{table}

As presented, reward is sparse; however, we leverage touch and flow to make the reward more dense.
If the touch sensor feels no force but flow exists, we know the agent should not hold or push in areas of no flow, which should correspond to the background.
We can safely supervise with reward 0 for all such pixels for both hold and push reward maps.
If the touch sensor feels a force, flow exists, and we have moved a new or existing part, then we can make the push reward dense.
We compute the L2-norm of the flow field and normalize by the max value.
If we moved a new part, these values give a dense target for the push map prediction.
If we moved an existing part, we scale the target dense push values by the \textit{existing part} reward of 0.5.

\subsection{Putting Everything Together: Act the Part}
\label{sec:full_pipeline}

We begin by training the interaction network using motion masks from the thresholded flow for history aggregation.
We then train our part network using the entire dataset of interactions to learn to infer motion masks.
At inference, we first predict and execute an action.
We infer motion masks and run history aggregation to output a segmentation mask at every timestep.
Further architecture and training details are provided in Appx. \ref{appx:architecture} and \ref{appx:training}.
\section{Evaluation}
\label{sec:evaluation}

Five Act the Part (AtP) models, trained with different seeds, are evaluated on 20 unseen instances from four seen categories (scissors,  knife, USB, safe) and 87 instances from five unseen categories (pliers, microwave, lighter, eyeglasses, and multilink).
The multilink objects have three links in a chain similar to eyeglasses.
Critically, all train instances have \textit{two links}; however, during testing, we evaluate on objects with \textit{two} and \textit{three links}.
See Appx. \ref{appx:assests} for information about the number of instances per category.

\begin{figure*}[t]
    \centering
    \includegraphics[width=40pc]{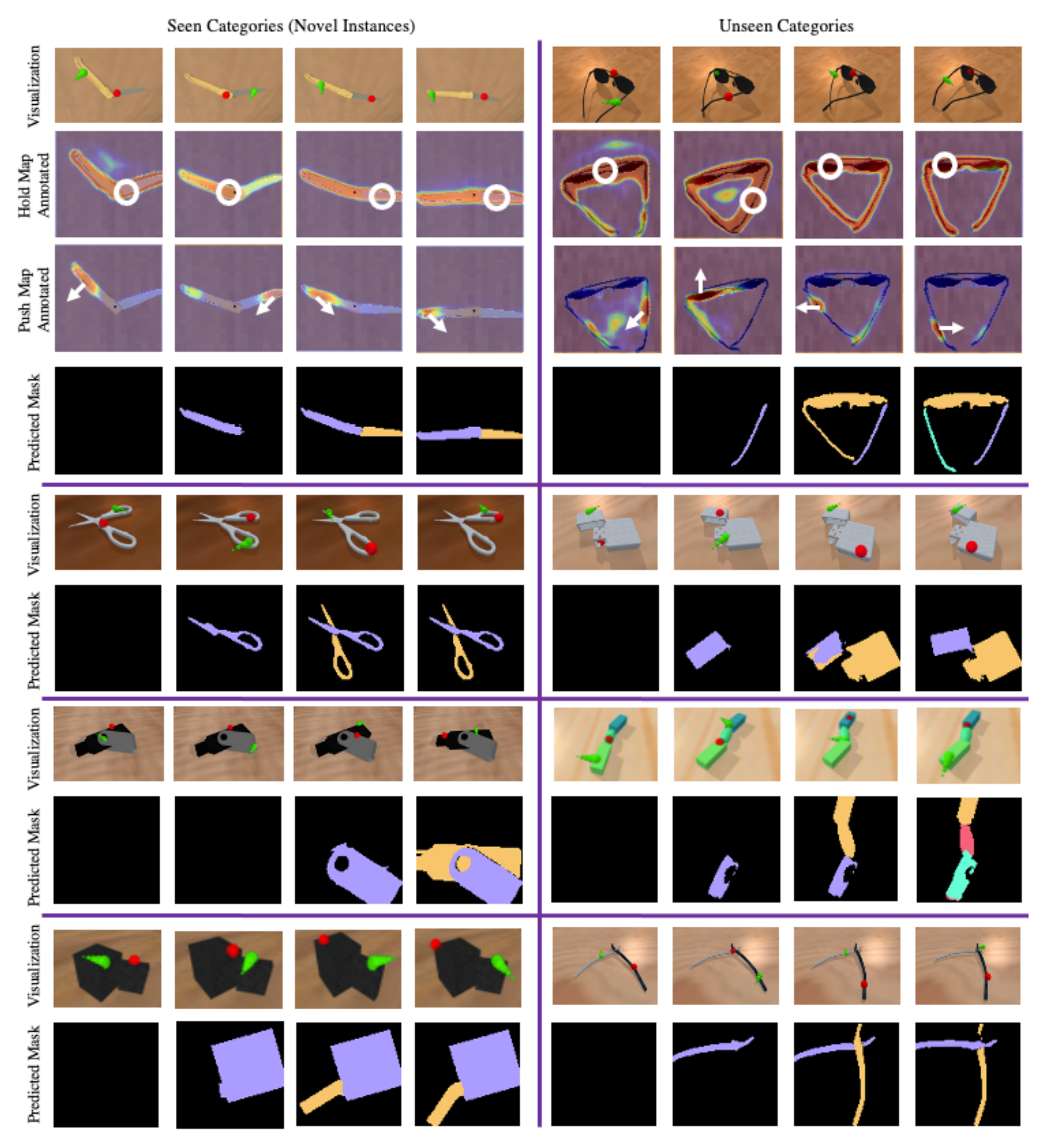}
    \caption{\textbf{Qualitative Results.}
    Our network learns a policy to interact with unseen objects and categories.
    While it is only trained on objects with two part, it also learns to reason about three part objects.
    Due to space limitation, only three interaction steps are shown in this figure.
    For more results, please refer to our project website: \href{https://atp.cs.columbia.edu}{https://atp.cs.columbia.edu}.}
    \label{fig:result}
\end{figure*}

To initialize instances, we uniformly sample start position, orientation, joint angles, and scale.
Dataset, test initialization, and pre-trained models will be released for reproducibility and benchmarking.

\subsection{Metrics and Points of Comparison}
For each test data point, we allow the agent to interact with the object five times.
We collect three perceptual metrics to evaluate performance on part discovery and segmentation.
Two additional metrics measure effectiveness of the actions for part discovery.
Let $\mathcal{G}$, $\mathcal{H}$ denote the sets of ground truth and predicted binary part masks respectively.

\mypara{Average Percentage Error (APE).}
To measures errors in number of parts discovered, we compute $|(|\mathcal{G}|-|\mathcal{H}|)|/|\mathcal{G}|$.

\mypara{Part-aware Intersection over Union (IoU).}    
We use Hungarian matching to solve for the maximal IoU bipartite match between $\mathcal{G}$ and $\mathcal{H}$.
Unmatched parts get IoU of 0.
Final IoU is determined by summing part IoUs and dividing by $\max(|\mathcal{G}|, |\mathcal{H}|)$.
The metric penalizes both errors in mask prediction and failure to discover masks (e.g. if one of two parts is discovered, maximum IoU is 50\%).

\mypara{Part-aware Hausdorff distance @ 95\% ($\mathbf{d_{H95}}$).}
We notice IoU is sensitive for thin structures.
For example, a small pixel shift in a thin rod can lead to IoU of 0.
To provide a better metric for these structures, we measure $d_{H95}$, which is a part-aware variant of a common metric in medical image segmentation \cite{hausdorff}.
The directed Hausdorff distance @ 95\% between some masks $G \in \mathcal{G}$ and $H \in \mathcal{H}$ is defined as $d^d_{H95}(G, H) = \underset{g \in G}{P_{95}} \min_{h \in H} ||g - h||_2 $
where $P_{95}$ gives the 95-th percentile value over pixel distances.
The metric is robust to a small number of outliers, which would otherwise dominate.
The symmetric measure is given as: 
$ d_{H95}(G, H) = \max (d^d_{H95}(G, H), d^d_{H95}(H, G))$.
We use Hungarian matching to find minimal $d_{H95}$ bipartite matches between $\mathcal{G}$ and $\mathcal{H}$.
If $|\mathcal{G}| \neq |\mathcal{H}|$, we compute the distance of unmatched parts against a matrix of ones at the image resolution.
Distances are summed and normalized by $\max(|\mathcal{G}|, |\mathcal{H}|)$.

\mypara{Effective Steps.}
A step is effective if the hold is on an object link, the push is on another link, and the action creates motion.

\mypara{Optimal Steps.}
An interaction is optimal if it is effective and a new part is discovered.
If all the parts have already been discovered, moving a \textit{single} existing part in the interaction is not penalized.

We compute the average of perceptual metrics for each category at every timestep over five models trained with different random seeds. 
Hence IoU, APE, and $d_{H95}$ yield mIoU, MAPE, and $\Bar{d}_{H95}$.
For evaluation in Tab. \ref{tab:exp}, we consider metrics after the fifth timestep.
Efficient and optimal action scores are averaged for each category over all timesteps (in contrast to being considered only at the fifth timestep).

\mypara{Baselines and Ablations.}  
We compare the AtP framework trained with and without touch reward, [Ours-Touch] and [Ours-NoTouch] respectively, with the following alternative approaches to study the efficacy of our interaction network.
All methods use the same part network trained from the full AtP rollouts:
\begin{itemize}[leftmargin=*]
\vspace{-2mm}
\item Act-Random: hold and push locations and the push direction are uniformly sampled from the action space.
\vspace{-2mm}
\item Act-NoHold: The agent applies a single push action every step. Single modes of interaction are widely used in interactive segmentation algorithms \cite{Fitzpatrick_2003, bjorkman_2010, van_hoof_2014, Pajarinen_2015, Pathak_2018, Eitel_2019}; however, this ablation differs from these works as push is learned.
\vspace{-2mm}
\item Act-NoPart: The interaction network does not take the part memory and considers each moved part as a new part for reward calculation.
\end{itemize}
\vspace{-2mm}
\noindent For the modified reward used to train the above networks see Appx. \ref{appx:ablated_versions}.
We also design two oracle algorithms using simulation state to provide performance upper bounds: 
\vspace{-2mm}
\begin{itemize}[leftmargin=*]
\item GT-Act: Optimal action based on ground truth state, but use of AtP part network for the mask inference. This is conceptually similar to \cite{Pillai_2014}, which uses expert actions for part segmentation.
\vspace{-2mm}
\item GT-Act-Mot: Optimal action based on ground truth state with motion masks from the ground truth flow. 
\end{itemize}

\begin{table*}
 \scriptsize  
  \centering
  \tabcolsep=0.04cm
  \begin{tabular}{l?c|c|c|c?c|c|c|c|c}         
%   \begin{tabular}{l|cccc|ccccc}         
  \toprule
                          & \multicolumn{4}{c?}{Train Categories} & \multicolumn{5}{c}{Test Categories} \\
    \midrule
           &  Scissors    & Knife          & USB             & Safe        & Pliers       & Microwave   & Lighter      & Eyeglasses  & Multilink \\
    Method                & $\downarrow$ / $\downarrow$ / $\uparrow$  & $\downarrow$ / $\downarrow$ / $\uparrow$  & $\downarrow$ / $\downarrow$ / $\uparrow$  & $\downarrow$ / $\downarrow$ / $\uparrow$  & $\downarrow$ / $\downarrow$ / $\uparrow$  & $\downarrow$ / $\downarrow$ / $\uparrow$  & $\downarrow$ / $\downarrow$ / $\uparrow$  & $\downarrow$ / $\downarrow$ / $\uparrow$  & $\downarrow$ / $\downarrow$ / $\uparrow$ \\\midrule

GT-Act & 0.01 / 4.3 / 78.0 & 0.02 / 4.5 / 81.6 & 0.00 / 5.7 / 82.7 & 0.02 / 2.1 / 89.7 & 0.01 / 4.3 / 78.4 & 0.03 / 2.4 / 87.8 & 0.00 / 3.3 / 88.0 & 0.03 / 7.4 / 64.6 & 0.10 / 7.2 / 75.3\\
GT-Act+Mot & 0.00 / 1.6 / 88.4 & 0.00 / 0.9 / 92.9 & 0.00 / 2.4 / 91.5 & 0.00 / 0.6 / 91.7 & 0.00 / 0.9 / 92.7 & 0.00 / 0.4 / 94.2 & 0.00 / 0.6 / 94.8 & 0.00 / 4.2 / 82.5 & 0.00 / 5.2 / 86.9\\
    \midrule 
Act-Random & 0.62 / 36.2 / 22.0 & 0.63 / 41.6 / 24.1 & 0.47 / 26.8 / 33.1 & 0.62 / 40.8 / 32.5 & 0.56 / 39.9 / 25.0 & 0.58 / 37.9 / 36.0 & 0.62 / 40.4 / 24.5 & 0.70 / 53.5 / 12.5 & 0.78 / 52.6 / 10.7\\

Act-NoHold & 0.46 / 34.4 / 28.5 & 0.43 / 35.5 / 35.2 & 0.40 / 30.0 / 33.2 & 0.38 / 32.1 / 42.7 & 0.41 / 35.4 / 30.5 & 0.41 / 31.3 / 40.7 & 0.45 / 39.6 / 34.3 & 0.40 / 41.6 / 19.6 & 0.53 / 48.0 / 19.9\\

Act-NoPart & 0.25 / 15.3 / 53.8 & 0.29 / 20.4 / 51.6 & 0.44 / 19.5 / 47.6 & 0.37 / 18.3 / 49.1 & 0.34 / 20.6 / 48.5 & 0.43 / 19.7 / 46.3 & 0.49 / 26.4 / 42.0 & 0.33 / 27.1 / 34.4 & 0.40 / 31.0 / 39.7\\
\midrule
Ours-NoTouch & 0.19 / 13.1 / 58.0 & \textbf{0.16} / 13.2 / \textbf{66.2} & 0.14 / 10.4 / 68.6 & \textbf{0.15} / \textbf{9.9} / \textbf{75.0} & 0.15 / 12.5 / 59.8 & \textbf{0.14} / 9.2 / 74.1 & \textbf{0.22} / \textbf{14.5} / \textbf{64.4} & 0.28 / 26.2 / 37.9 & 0.25 / 24.6 / 46.6\\

Ours-Touch & \textbf{0.10} / \textbf{8.5} / \textbf{65.6} & \textbf{0.16} / \textbf{12.2} / 65.9 & \textbf{0.09} / \textbf{8.3} / \textbf{75.3} & 0.17 / 10.1 / 74.2 & \textbf{0.13} / \textbf{9.7} / \textbf{64.9} & \textbf{0.14} / \textbf{8.3} / \textbf{75.4} & 0.25 / 15.1 / 62.8 & \textbf{0.24} / \textbf{21.8} / \textbf{43.0} & \textbf{0.22} / \textbf{20.0} / \textbf{54.7}\\

\bottomrule
  \end{tabular}
  \caption{ \textbf{Perception performance.} MAPE [frac] / $\overline{d}_{H95}$ [pixels] / mIoU [\%]. Image resolution is $90 \times 90$. Numbers are evaluated after the fifth interaction. Numbers are averaged over five models trained with different seeds.}\label{tab:exp}
\end{table*}

\subsection{Benchmark Results}  

To validate the major design decisions, we run a series of quantitative experiments in simulation to compare different algorithms.
We also provide qualitative results in Fig. \ref{fig:result}.
In Sec. \ref{sec:real}, we evaluate our model on real world data.

\begin{figure}[t]
\includegraphics[width=\linewidth]{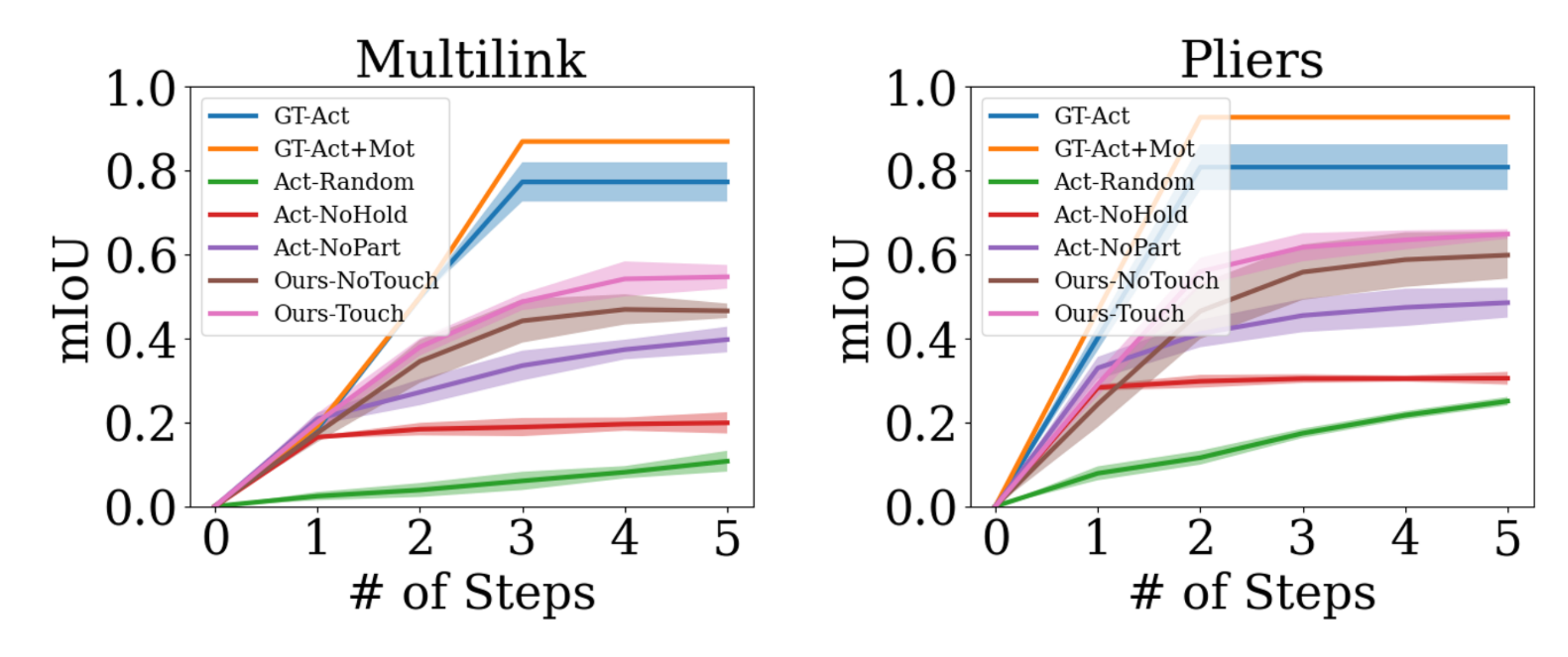}
\caption{\textbf{IoU w.r.t. Interaction Steps.}
Results on two unseen object categories show our methods (pink and brown) approach the oracle baseline over time.}
\label{fig:iousteps}
\end{figure}

\mypara{Does Interaction Help Part Discovery?} 
First we want to validate if AtP learns effective interaction strategies for part discovery by accumulating information over time.
To evaluate, we plot the part mIoU w.r.t. interaction steps in Fig. \ref{fig:iousteps}.  
As expected, the upper bounds peaks at 2 and 3 steps for pliers and multilink, respectively.
While other algorithms' performance saturate quickly with one or two interactions, [Ours-Touch] and [Ours-NoTouch] are able to improve with more interactions. 
These plots indicate that while the learned interaction strategies may not be optimal (compared to upper bounds using ground truth state), they are informative for discovering new parts of the object and self-correct errors over time. 
Results from other categories are presented in Appx. \ref{appx:additional_results}, where we see all AtP curves approach the upper bounds.

\begin{figure}[t]
\includegraphics[width=0.98\linewidth]{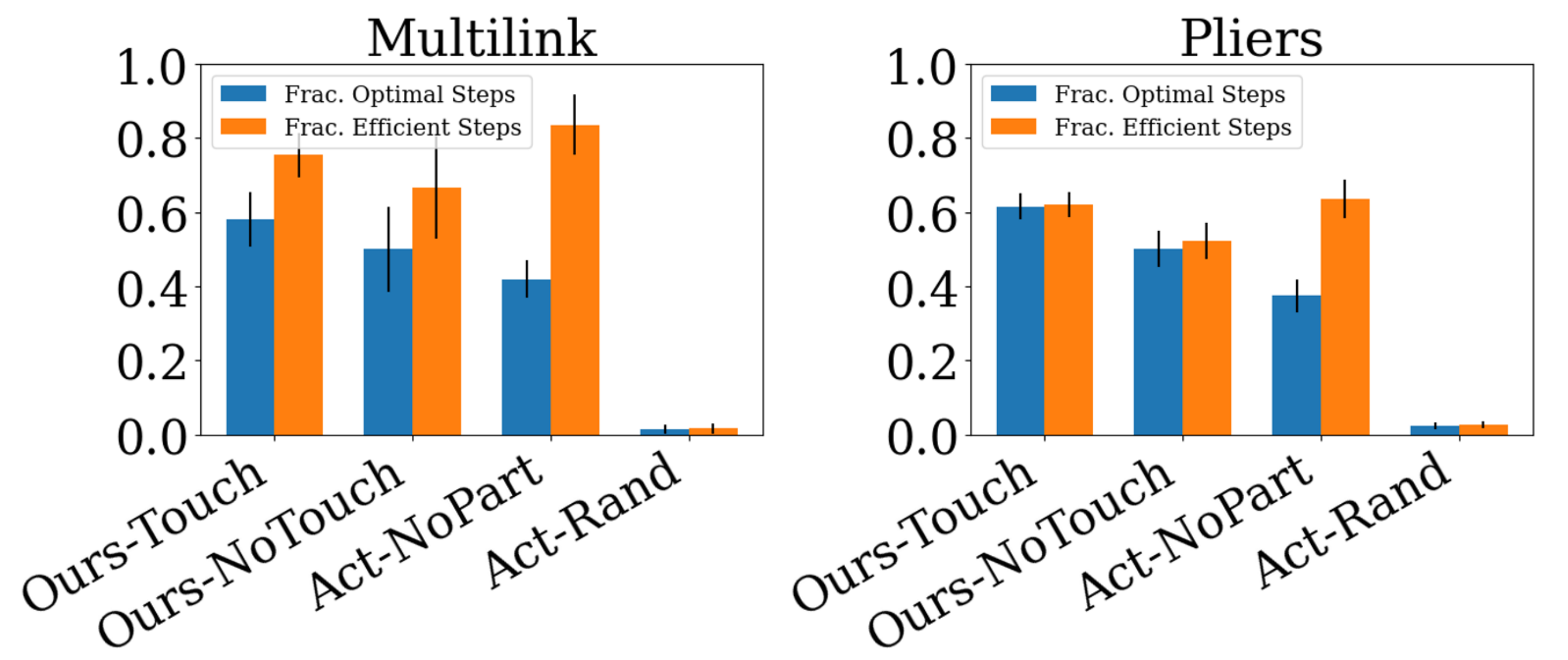}
\caption{\textbf{Effective and Optimal Actions.}
Our method learns an efficient policy that chooses optimal actions (i.e., actions that discover new parts) more frequently than other approaches.} 
\label{fig:actions}
\end{figure}

\mypara{Does Part Prediction Help with Action Selection?} 
Our interaction network takes the current belief of the part segmentation as input and obtains reward for new part discovery.
We hope this design would encourage the algorithm to focus on selecting actions that provide information gain (e.g., push new parts to discover them). 
To validate this design, we compare AtP to an ablated version, [Act-NoPart], which is not mask-aware.
Interestingly, this model performs efficient actions at roughly the same rate as [Ours-Touch] (Fig. \ref{fig:actions}); however, [Ours-Touch] is better at finding optimal actions (resulting in new part discovery).
Histograms for all other categories are presented in Appx. \ref{appx:additional_results} and corroborate these findings.
This result is also supported in Tab. \ref{tab:exp}, which shows degradation on all perceptual metrics when part-awareness is not exploited.

\mypara{Is Holding Necessary?} 
In contrast to a vast majority of prior work that use simple push actions, our algorithm uses bimanual actions for object interaction (i.e., simultaneous hold and push).
Our hypothesis is that such actions give the system a better chance at disentangling motion between different moving parts and therefore aid part discovery. 
To validate this hypothesis, we compare our algorithm with an agent that performs only push actions [Act-NoHold]. 
The result in Tab. \ref{tab:exp} shows that without the hold action the system performance is much worse at part segmentation.
[Act-NoHold] has has trouble discovering more than one object part, since the whole object is likely to be moved during interaction.
Furthermore, this result suggests more complex perceptual modules are necessary to get push-only policies to achieve competitive performance at this task.
While this is an interesting direction, disentangling the motion of many moving parts is non-trivial and out of scope for this paper.

\mypara{Does Touch Feedback Help?} 
In this experiment, we want to evaluate the effect of touch feedback.
Looking at Tab. \ref{tab:exp}, we see that [Ours-Touch] outperforms [Ours-NoTouch] in most categories.
A similar trend is noticeable when looking at action performance in Figs. \ref{fig:iousteps} and \ref{fig:actions}.
We conjecture this is due to the benefit of using touch signal to define more specific reward cases and to make reward more dense, which is ultimately reflected in the final system performance.
However, we are still able to learn helpful interaction strategies even without touch.

\mypara{Generalization to Unseen Objects and Categories.}
Our algorithm does not make category-level assumptions, therefore the same policy and perception model should work for unseen object categories with different kinematic structures. 
More specifically, we wish to probe generalization capabilities of our model to unseen instances from seen categories and novel categories.

The algorithm's generalizablity is supported by results in Tab. \ref{tab:exp}, where mIoU, MAPE, and $\overline{d}_{H95}$ are comparable for train versus test categories.
Performance on eyeglasses is slightly worse, however, still impressive as our model is only trained on instances with two links. 
Furthermore, for eyeglasses, MAPE value falls under 0.33, suggesting the model finds the three parts in most cases. 
IoU performance on the multilink category is better than on eyeglasses; however, MAPE is comparable, suggesting that eyeglasses are particularly challenging for reasons other than having three links.
These results support that our method learns to interact intelligently and reason about motion in spite of differing shape, texture, or structure in the test objects.

\subsection{Real World Results} \label{sec:real}

In these experiments, we want to validate [Ours-Touch] performance on real world data. Since our algorithm does not need prior knowledge of objects or special sensory input during inference, we can \textit{directly} test our learned model on real world RGB images of unseen categories taken by smartphone cameras.

To build a pipeline that demonstrates the viability of our model on real world data,
a camera is positioned over an articulated object and an image is captured.
Our trained model runs interaction inference, predicts hold and push actions, and provides a human operator with instructions on what to execute.
A next frame image is sent back to the model, at which point it runs the part network, history aggregation, and another round of interaction inference.
More details on the procedure can be found in Appx. \ref{appx:real_world}.

\begin{figure} 
  \includegraphics[width=\linewidth]{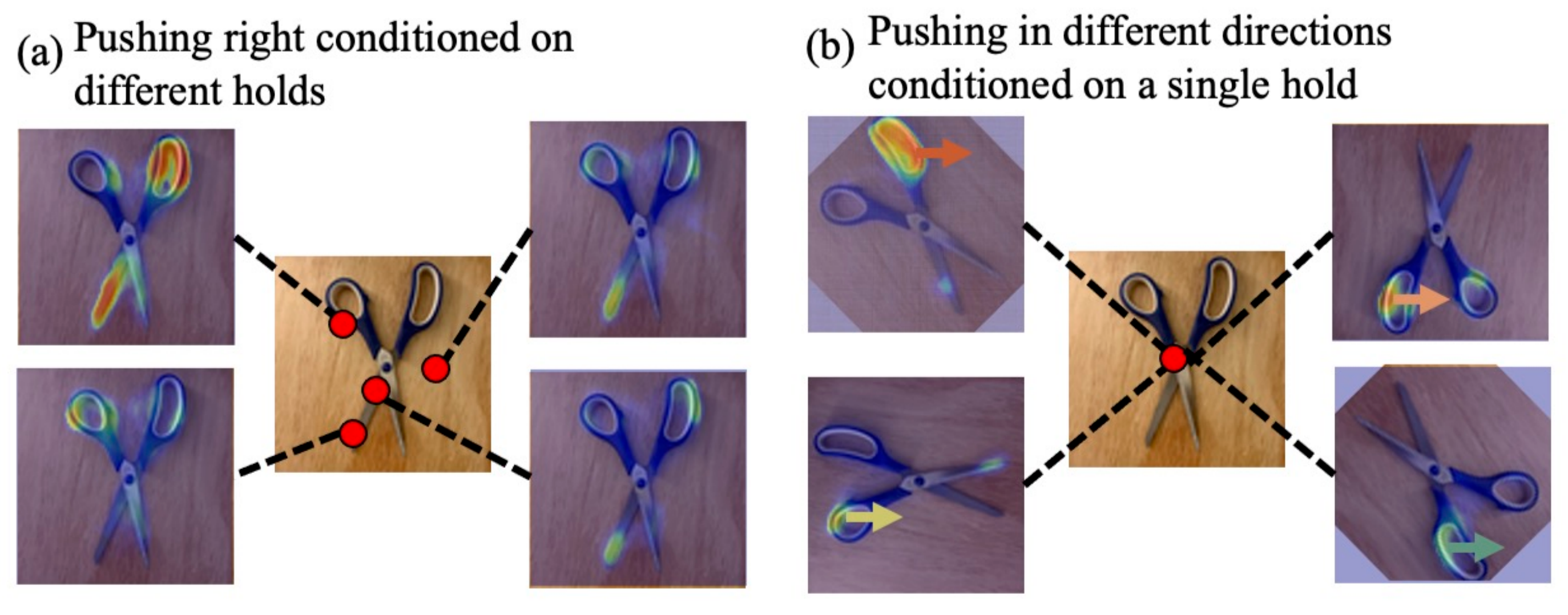}
  \caption{\textbf{Conditional Action on Real Images.} (a) Varying the hold location, we observe the model is able to reason where to push right. (b) Fixing the hold location, we observe the model reasons about a good direction to push (i.e. top left).} 
  \label{fig:probe}
\end{figure}

\mypara{Conditional Action Reasoning.}
We visualize the conditional action inference result from the interaction network on real world images. 
Fig. \ref{fig:probe} shows two types of visualizations. 
In example (a), we pick various hold positions and analyze the ``push right" reward prediction maps (recall: pushing is conditioned on holding).
We notice that the affordance prediction switches between the links depending on the hold location, which indicates the network's understanding about the object structure. 
When hold is placed in free space or between the links, the push reward predictions are not confident about pushing anywhere.
These results suggest that our model is able to disentangle push predictions from its own hold predictions, thereby demonstrating a form of conditional reasoning.

In example (b), we further probe the model's reasoning about the push direction by visualizing different push maps for the same holding position.
Among all directions, the network infers the highest score on the top-left rotation, which would push the scissors open.   
The result suggests that the algorithm is able to pick a push direction that would lead to informative motion, when reasoning over many options. 

\begin{figure} 
 \includegraphics[width=\linewidth]{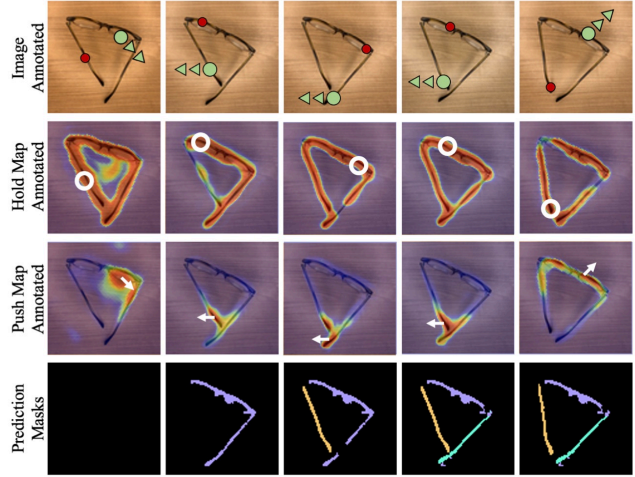}
    \caption{\textbf{Interaction Experiment.} Images were captured and sent to AtP, where hold and push are predicted. A human agent executes actions following the instructions. From pairs of images, masks are recovered and aggregated by AtP, and the next instruction is given. This is especially interesting because the lighting, shadows, and artifacts of the images taken with a phone are different from our simulated environment (i.e. sim2real gap).} 
    \label{fig:real}
\end{figure}

\mypara{Interaction Experiment.}
Next, we evaluate both perception and interaction networks together with the real world physical interactions. To validate performance independent of robot execution accuracy, a human is instructed to execute the actions. 
Fig. \ref{fig:real} shows the predicted actions, affordances and final object part masks discovered by the algorithm.
Without any fine-tuning, the algorithm shows promising results on inferring interaction strategies and reasoning about the observed motion for part discovery.
Please refer to Appx. \ref{appx:additional_results} for more real world experiment results and failure case analysis.
Note that no quantitative analysis can be made for this experiment as there is no ground truth part segmentation for these images.

\section{Conclusion and Future Work} 

We present Act the Part (AtP) to take visual observations of articulated objects, interact strategically, and output part segmentation masks.
Our experiments suggest: (1) AtP is able to learn efficient strategies for isolating and discovering parts.
(2) AtP generalizes to novel categories of objects with unknown and unseen number of links---in simulation and in the real world.
(3) Our model demonstrates conditional reasoning about how to push based on arbitrary hold locations.
We see broad scope for future work including extensions to 3D part segmentation and singular frameworks for rigid, articulated, and deformable object understanding.
We hope this paper will inspire others in the vision and robotics communities to investigate perception and interaction in tandem.
% \vspace{3mm}

\small{
\mypara{Acknowledgements.}
Thank you Shubham Agrawal, Jessie Chapman, Cheng Chi, the Gadres, Bilkit Githinji, Huy Ha, Gabriel Ilharco Magalhães, Sarah Pratt, Fiadh Sheeran, Mitchell Wortsman, and Zhenjia Xu for valuable conversations.
This work was supported in part by the Amazon Research Award and NSF CMMI-2037101.
}
\normalsize

% \newpage 
{\small
\bibliographystyle{ieee_fullname}
\bibliography{egbib}
}
\newpage
\appendix
\begin{figure*}[t]
\includegraphics[width=40pc]{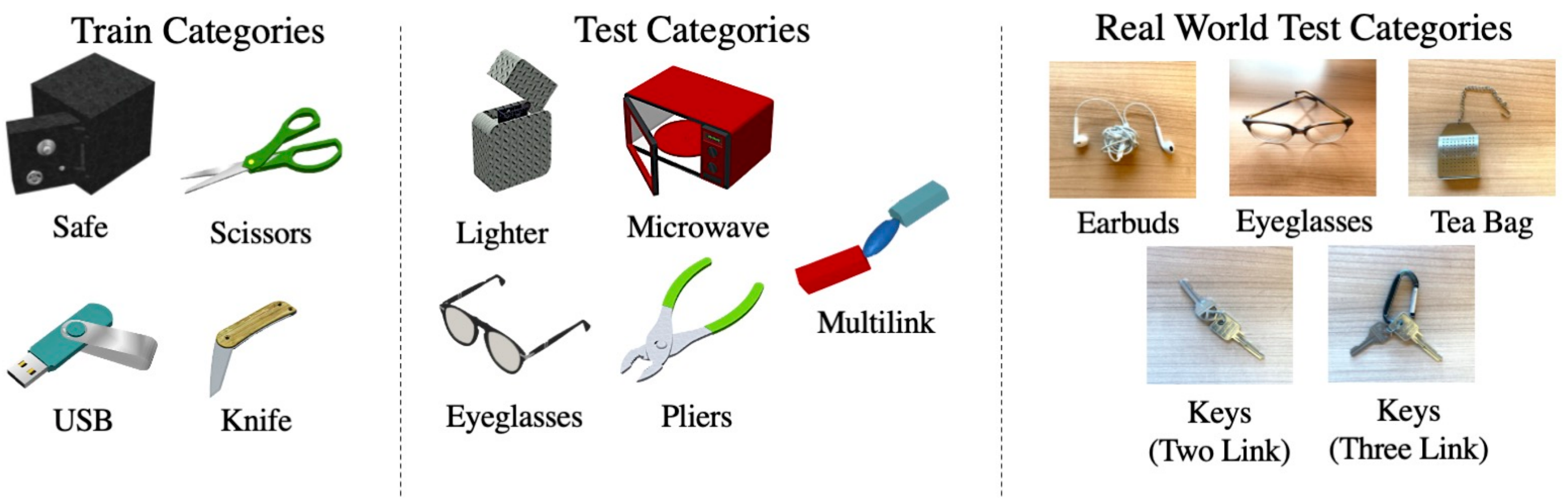}
\caption{\textbf{Categories.} Sample of instances from train and test categories (synthetic and real world). Note during test time, unseen instances of train categories are also evaluated.}
\label{fig:categories}
\end{figure*}

\section{Data Generation}

In this section we give more details about our data generation pipeline for Act the Part (AtP).

\subsection{Assets}
\label{appx:assests}

For object asset data, we use instances from eight categories from the Partnet-Mobility \cite{chang_2015, Mo_2019, Xiang_2020} dataset and a ninth multilink category of our creation---configured with three links in a chain.
Examples from each category are shown in Fig. \ref{fig:categories}.
Number of instances for train splits, test splits, and the number of initial poses per category for testing can be found in Tab. \ref{tab:counts}.

\mypara{Partnet-Mobility.} 
Object categories were selected for realism in table-top environments and provide opportunity for reasonable exploration of the hold and push action pair.
For this work, we also consider top-down views, which are commonly used in robot bin clearing tasks.
Notably, not enough categories with prismatic joint (e.g. furniture) fit these parameters.
We leave tackling such objects to future work.
We additionally filtered instances with missing 3D object meshes or unstable physical dynamics.

\mypara{Multilink.}
Our multilink instances are composed of three links, with an ellipsoid flanked by two prisms.
Each link is assigned a random color by sampling R, G, and B values uniformly with replacement.
Joint angles take values in the range $[-5\pi/18, 5\pi/18]$ radians.
We also introduce tooling to create arbitrary multilink instances with mesh primitives, which can be used in future work and applications.
Note: procedurally generated multilink assets follow Partnet-Mobility conventions.
Code for generating the multilink objects will be available online.

\begin{table}
 \scriptsize  
  \centering
  \tabcolsep=.18cm
  \begin{tabular}{lccc}         
  \toprule
         Category & \# Train Instances & \# Test Instances & \# Test Pose Initializations\\
    \midrule
    Scissors & 37 & 9 & 90\\
    Knife & 18 & 4 & 40\\
    USB & 15 & 3 & 30\\
    Safe & 18 & 4 & 40\\
    Pliers & - & 24 & 48\\
    Microwave & - & 9 & 18\\
    Lighter & - & 11 & 22\\
    Eyeglasses & - & 33 & 66\\
    Multilink & - & 10 & 20\\\midrule
    Total & 88 & 107 & 374\\
    \bottomrule
  \end{tabular}
  \caption{ \textbf{Instance Counts.} Details about the number of assets and testing pose initializations.}\label{tab:counts}
\end{table}

\subsection{Physics Simulator}
\label{appx:physics_simulator}
Our Simulation platform is based on Pybullet~\cite{coumans2016pybullet}, a state-of-the-art physics simulator and wrapper for the Bullet Physics Library.
Pybullet is widely used in vision, robotics, and reinforcement learning.

\mypara{Image Backgrounds.}
We use wood textured backgrounds sourced from Kaggle\footnote{\href{https://www.kaggle.com/edhenrivi/wood-samples}{kaggle.com/edhenrivi/wood-samples}}.
The original data is scraped from the Wood Database\footnote{\href{https://www.wood-database.com}{wood-database.com}}.
We use 421 backgrounds from the dataset and partition into three sets: 141 instances for training, 140 for testing unseen instances from seen categories, and 140 for testing unseen categories.
When initializing the Pybullet environment, depending on the train or evaluation setting, a background is drawn uniformly at random.
For the testing, this is done \textit{once} and initialization conditions are frozen to ensure a fixed benchmark.
All initial poses and corresponding backgrounds will be released in our project code release.

\mypara{Simulating Touch Feedback.}
We simulate sheer force touch feedback in our hold gripper.
The touch feedback is measured by sensing constraint forces on the hold gripper.
These are non-zero when hold and push are both on the same object.
The existence of a constraint force indicates that the gripper must hold more forcefully to keep a point on the object fixed.
While we can obtain rich signal including direction and magnitude of the forces, we limit our agent to a binary signal, which is more reliable in real world settings.

\section{History Aggregation}
\label{appx:history_aggregation}

\mypara{Full Part memory.}
It is possible that all channels in part memory $V$ are full when we want to allocate a new part $M_{t+1}$.
In these cases, we assign to the channel $c$ with the largest overlap with $M_t$: $V^c \gets (V^c - (M_t \cap V^c)) \cup M_{t+1}$.
This assignment potentially entangles two masks.
All such cases are classified as \textit{entangled parts}.

\mypara{Perspectives on History Aggregation.}
The algorithm can be viewed as maintaining $V$ as a hidden state.
However, we opt to use our history aggregation module instead of an RNN.
Training an RNN would require pseudo-labels, which realistically come from this algorithm or additional ground truth supervision.

\mypara{Implementation Details.}
$V$ is implemented as a 3D tensor.
In this work we deal with object with at most three links.
However, in practice we use five channel tensor to relax the assumption that we can deal with only three parts.
A priority queue maintains order of the most recently allocated and modified channels.
This allows us to turn $V$ into a single segmentation mask $\mathcal{M}$, by layering channels in an occlusion aware order.

\section{Reward}
\label{appx:reward}

\mypara{Full Model with Touch.}
If no flow is observed in the scene, it implies the push is not proper, while not necessarily implying anything about the hold action.
Hence, the push reward is 0 and no hold reward is back-propagated.

If flow exists, it implies the push is on the object; however, there is no guarantee of helpful motion.
To better identify informative motion, we use touch feedback.
(1) When hold and push are both on the object, the hold gripper will feel sheer force (thresholded at 0 to give binary signal), indicating the need to hold harder to keep a point fixed.
In this case,  (1a) we provide reward 1 to both hold and push pixels if a new part is discovered, and (1b) 0.5 if an existing part is moved.
(2) If the hold gripper feels no force and there is motion, it is clear the push action created motion without anything being pinned. The hold reward is 0 and no push reward is back-propagated.
(3) If the agent pushes a previously discovered part along with another undiscovered part, motions are entangled.
Here, we penalize hold with reward 0 and do not update push network, as pushing is conditioned on the hold.

When supervising push affordances, we also enforce reward 0 for the hold pixel, which should teach the agent not to push where it holds.

\mypara{Other Models.}
\label{appx:ablated_versions}
We show reward for [Act-NoHold], [Act-NoPart], and [Ours-NoTouch] in Tabs. \ref{tab:act_no_hold}, \ref{tab:act_no_part}, and \ref{tab:act_no_touch} respectively.

\begin{table}
 \scriptsize 
  \centering
  \tabcolsep=0.72cm
  \begin{tabular}{ccc}         
  \toprule
         Optical Flow  & Part Memory & Push Reward\\
    \midrule
    x & - & 0\\
    \checkmark & New part & 1 \\
    \checkmark & Existing part & .5 \\
    \checkmark & Entangled part & 0 \\
    \bottomrule
  \end{tabular}
  \caption{ \textbf{Act-NoHold Reward.} Reward cases related to holding are removed.}
  \label{tab:act_no_hold}
\end{table}

\begin{table}
 \scriptsize  
  \centering
  \tabcolsep=0.35cm
  \begin{tabular}{cccc}         
  \toprule
         Optical Flow & Touch Sensor & Hold Reward & Push Reward\\
    \midrule
    x & 1/0 & N/A & 0\\
    \checkmark & 1 & 1 & 1 \\
    \checkmark & 0 & 0 & N/A \\
    \bottomrule
  \end{tabular}
  \caption{ \textbf{Act-NoPart.} Reward related to part-awareness is removed.}
  \label{tab:act_no_part}
\end{table}

\begin{table}
 \scriptsize  
  \centering
  \tabcolsep=0.35cm
  \begin{tabular}{cccc}         
  \toprule
         Optical Flow & Part Memory & Hold Reward & Push Reward\\
    \midrule
    x & - & N/A & 0\\
    \checkmark & New part & 1 & 1 \\
    \checkmark & Existing part & .5 & .5 \\
    \checkmark & Entangled part & 0 & N/A \\
    \bottomrule
  \end{tabular}
  \caption{ \textbf{Ours-NoTouch.} Touch sensor is removed.}
  \label{tab:act_no_touch}
\end{table}

\section{Architecture}
\label{appx:architecture}

\mypara{Interaction Network.}
We use a ResNet18 \cite{He2016} backbone.
The conv1 layer is modified to change the number of input channels from three to nine.
No pretrained weights are used.
The nine channels are justified as follows.
The first three channels are used for an RGB image.
The next five channels are used for the part memory $V$.
The final channel is a hold channel, where an encoding of the hold location is passed when extracting the push map.
When computing the hold map, this channel is filled with placeholder zeros.

We have two decoder heads branching off of the shared encoder, one for holding and the second for pushing.
These heads are wired using residual connections similar to the U-Net architecture \cite{Ronneberger2015}.
We now describe a single upsampling block.
Features are bilinearly upsampled by a factor of 2.
A single conv layer is applied to reduce the number of channels by a factor of two.
The result is concatenated with intermediate features of the same resolution from the backbone pass.
The resulting volume is passed through two residual blocks, following the pytorch ResNet implementation.
Each upsampling head is composed of four upsampling blocks.
The output of these heads is a 1-channel map of logits used to predict reward for each input pixel.
The first head is used to predict hold rewards and the second to predict push rewards, as discussed in Sec. \ref{sec:interaction}.

For pushing, we want to simplify learning so the network only has to reason about pushing in a single direction (in our case right).
To accomplish this we take eight rotations of our input volume, every 45$\degree$.
To avoid data loss from rotations, we edge pad images (replicating edge values outward), before rotation and take a $128 \times 128$ center crop. 128 is sufficient as it preserve the diagonal of the original $90 \times 90$ image.

For the full forward pass, we first extract features from the image and part memory.
We then use the hold decoder to predict a hold map and sample to get the hold location.
The location is encoded as a gaussian.
The encoded hold replaces the channel of zeros in the input volume.
Features are again extracted by the backbone from eight rotations of the input.
The push decoder is used to get the push maps, where we can then choose the push direction and sample a location.

\mypara{Part Network.}
We use a ResNet18 backbone modified to take eight channel inputs.
The channels are filled with RGB images at timestep $t$ and $t+1$.
The last two channels are populated with gaussian encodings for the hold and the push.
Again, no pretrained weights are used.

The decoder heads are architecturally the same as in the interaction network.
They produce logit predictions for each pixel in the motion mask.

\section{Training}
\label{appx:training}

\mypara{Policy Rollouts for Data Collection.}
During each iteration of training, we rollout seven timesteps for each of the 88 train object instances from the scissors, knife, USB, and safe Part-Net Mobility categories.
The data and the corresponding reward is saved in a buffer. 
Our buffer holds the last $\beta$ iterations of data, whose distribution changes slowly as the model interactions improve.
We empirically find $\beta=10$ is a suitable rolling window to give good training set performance.

In total, the rollouts produce 73,920 interactions, from different stages in training.
This corresponds to 73,920 frame pairs, actions, and ground truth flow transitions (consistent forwards and backwards). 
Flow is thresholded at zero to produce the motion mask ground truth necessary for supervising the part network.

\mypara{Augmentations.}
We apply color jitter using pytorch APIs: brightness=0.3, contrast=0.4, saturation=0.3, hue=0.2.
Additionally we randomly set images to grayscale with probability 0.2.
After applying these augmentations, we normalize image RGB values using standard ImageNet mean and standard deviation.
For the part network, we sample different augmentations for images at timesteps $t$ and $t+1$.
For the part memory, we randomly swap channels to encourages invariance to the channel order.

\mypara{Training Details.}
The interaction and part networks are trained using SGD with momentum $0.9$, learning rate $\num{1e-3}$, weight decay $\num{5e-4}$, cosine-annealing schedule with t-max=120, batch size 64, for 120 epochs.
Hold map, push map, and part predictions are all supervised with binary cross entropy with logit loss.
Each network trains on a single GeForce RTX 2080 11 GB card in less than 12 hours.
For CPU rendering in the parallel rollouts, we make use of 48 Intel Xeon Gold 6226 (2.70GHz) cores.

\section{Real World Experimental Setup}
\label{appx:real_world}
\begin{figure}[t]
\includegraphics[width=\linewidth]{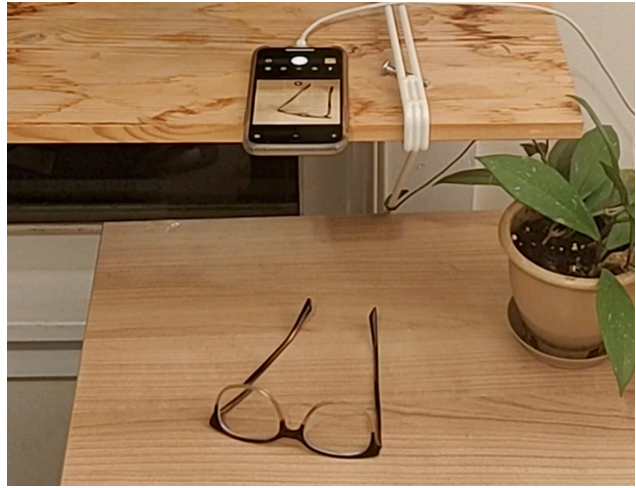}
\caption{\textbf{Real World Setup.} A simple configuration for taking picture of the object to send to a laptop running our model.}
\label{fig:setup}
\end{figure}

\begin{figure}[t]
\includegraphics[width=\linewidth]{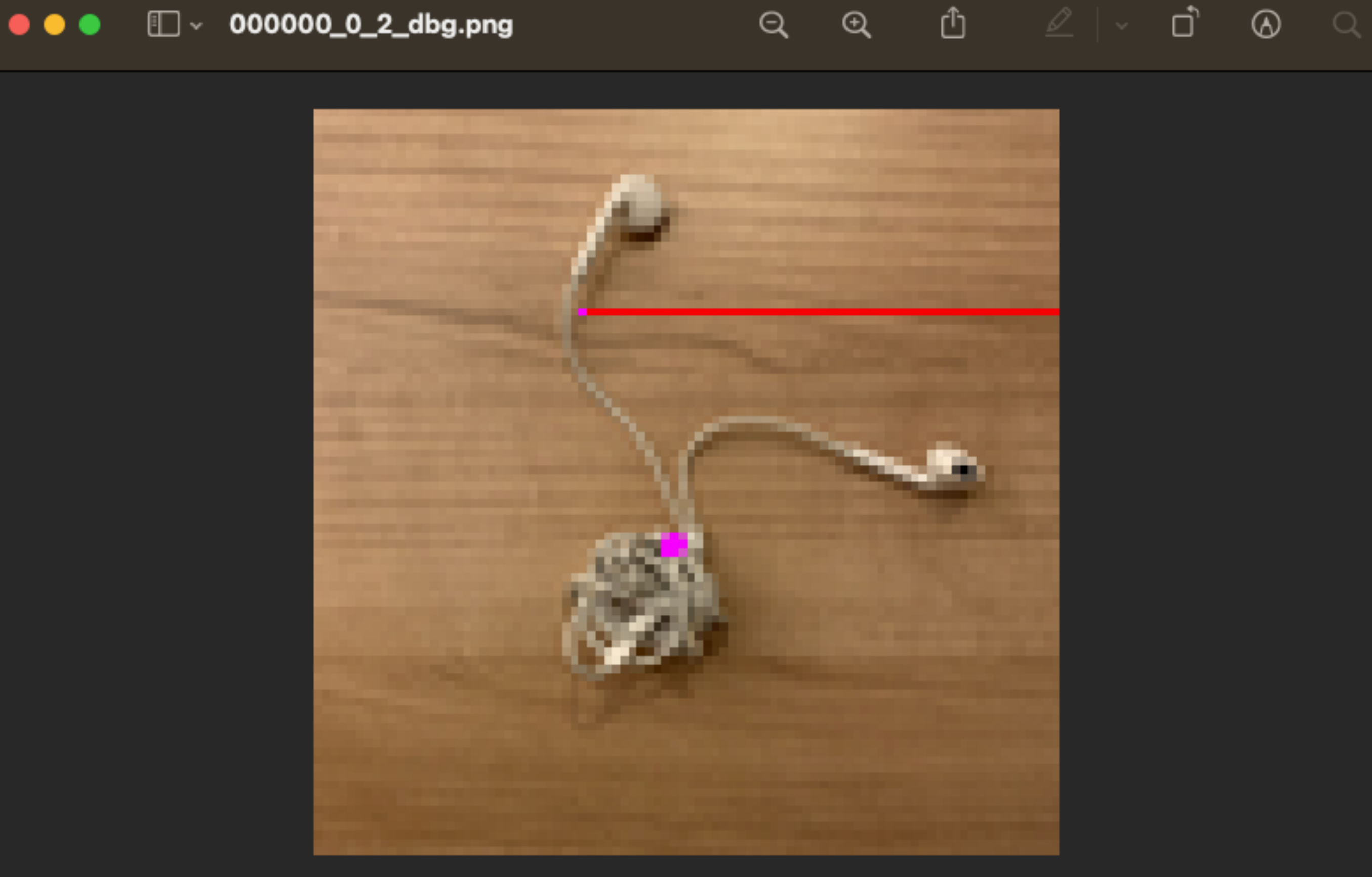}
\caption{\textbf{Instructions.} Our model indicates the action for a human operator to execute. The magenta dot indicates the location to hold. The red line indicates the start location and direction of the push. GUI is preserved to give context for what one would see when conducting real world experiments.}
\label{fig:gui}
\end{figure}

To conduct experiments in the real world we follow the following procedure.
\begin{itemize}[leftmargin=*]
\vspace{-2mm}
\item Take a picture of the object using an iPhone (setup shown in Fig. \ref{fig:setup}).
\vspace{-2mm}
\item Send picture to a laptop.
\vspace{-2mm}
\item Resize image to $90 \times 90$.
\vspace{-2mm}
\item Run AtP Interaction Network inference and visualize the selected hold location and push location (displayed in Fig. \ref{fig:gui}).
\vspace{-2mm}
\item Have human execute the action.
\vspace{-2mm}
\item Snap another picture and send to the computer.
\vspace{-2mm}
\item Run AtP Part Network inference to recover the part mask for this timestep.
\vspace{-2mm}
\item Repeat until last timestep at which point stop.
\vspace{-2mm}
\end{itemize}

\section{Additional Results}
\label{appx:additional_results}
Due to space limitations, we only presented interaction step and action results for microwave and eyeglasses categories.
Furthermore, we only showed real world results on eyeglasses.
Here we provide additional qualitative and quantitative results.

\begin{figure}[t]
\includegraphics[width=\linewidth]{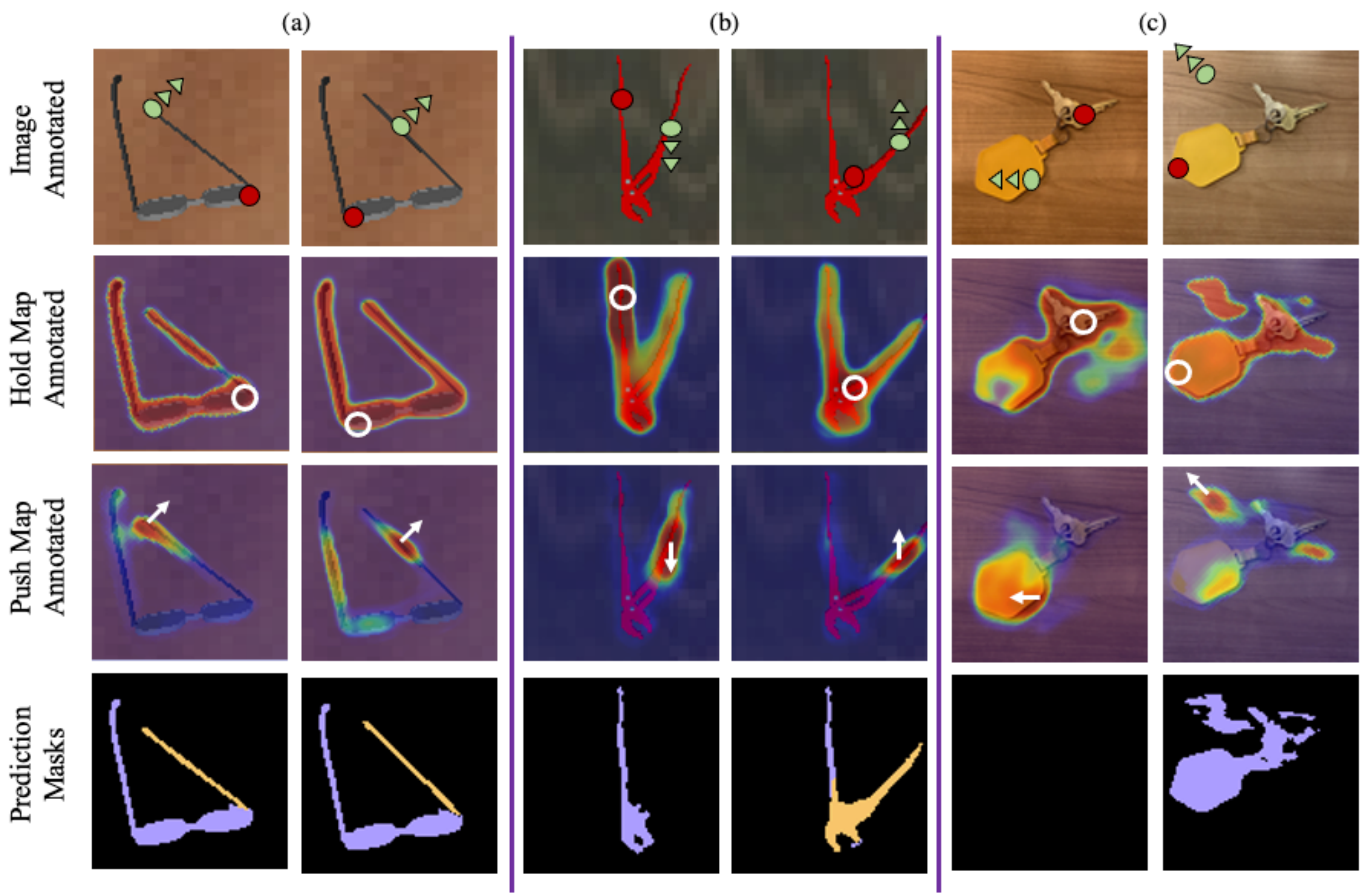}
\caption{\textbf{Failure Modes.} (a) On three link objects our model sometimes struggles to split parts that have been grouped together in the part memory. We conjecture that this is due to the fact that we train on only two link objects. Because there are not many instances of having to split masks, the network might already think it has discovered all the parts. (b) We call attention to the erroneous (over) segmentation of the moved part. We notice that this is a common failure mode of our part network. (c) This is another case of erroneous segmentation, this time in a real world example. Notice that the camera gain is different between the two frames. Because segmentation is poor, subsequent action selection (i.e. the push in the top right) can suffer.}
\label{fig:failure}
\end{figure}

\mypara{Failure Analysis.}
See Fig. \ref{fig:failure}, where comments on the failures are provided in the caption.

\begin{figure*}[t]
\includegraphics[width=40pc]{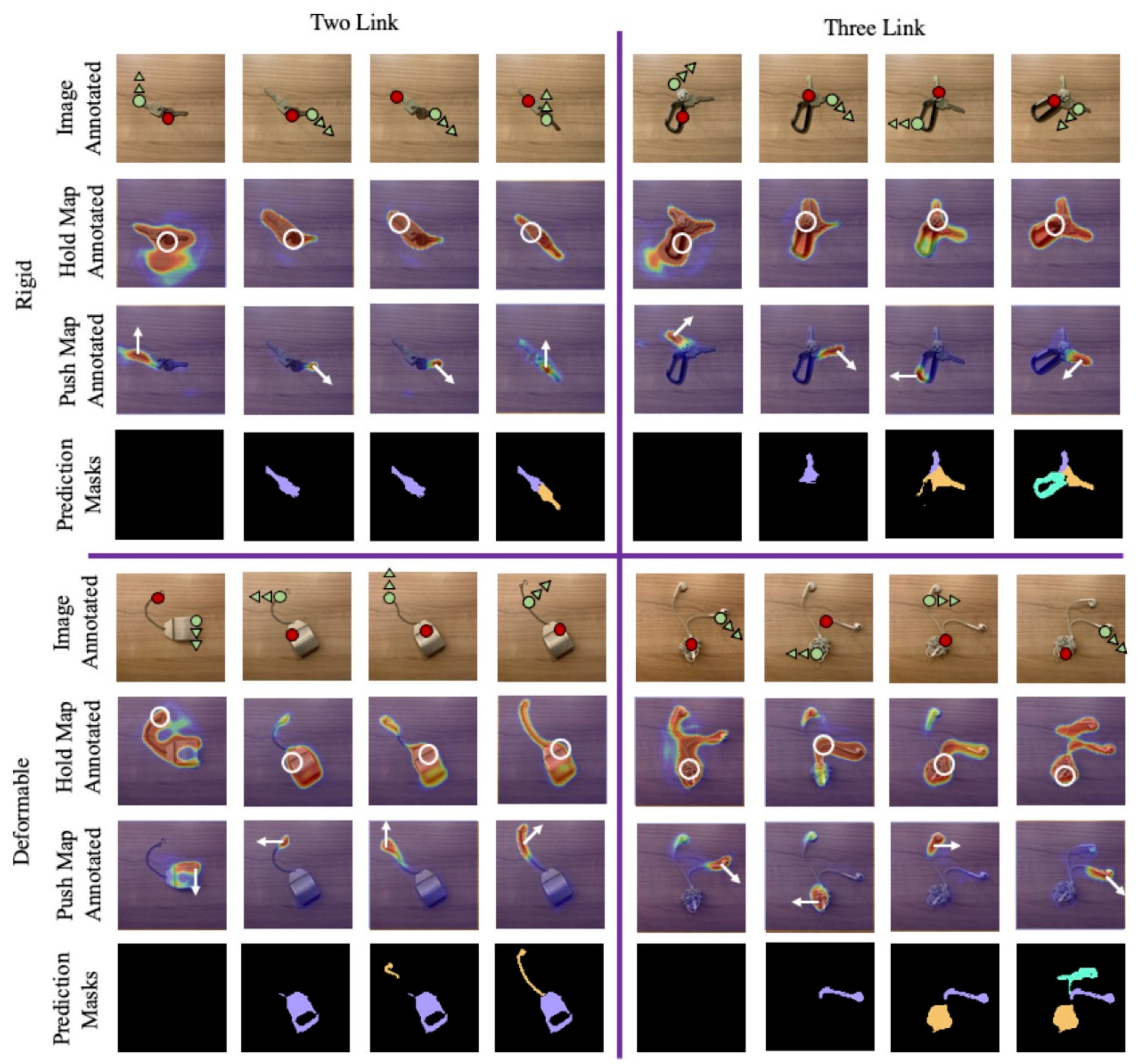}
\caption{\textbf{Real World Results.} More results on real world objects.}
\label{fig:more_real_world}
\end{figure*}

\mypara{Real World Results.}
We show results on four additional unseen categories: keys (two link, rigid), keys (three link, rigid), tea bag (two link, deformable), and earbuds (three link, deformable) in Fig. \ref{fig:more_real_world}.
Surprisingly our model works on these instances, even in the presence of deformable parts, which are not seen during training.

\begin{figure*}[t]
\includegraphics[width=40pc]{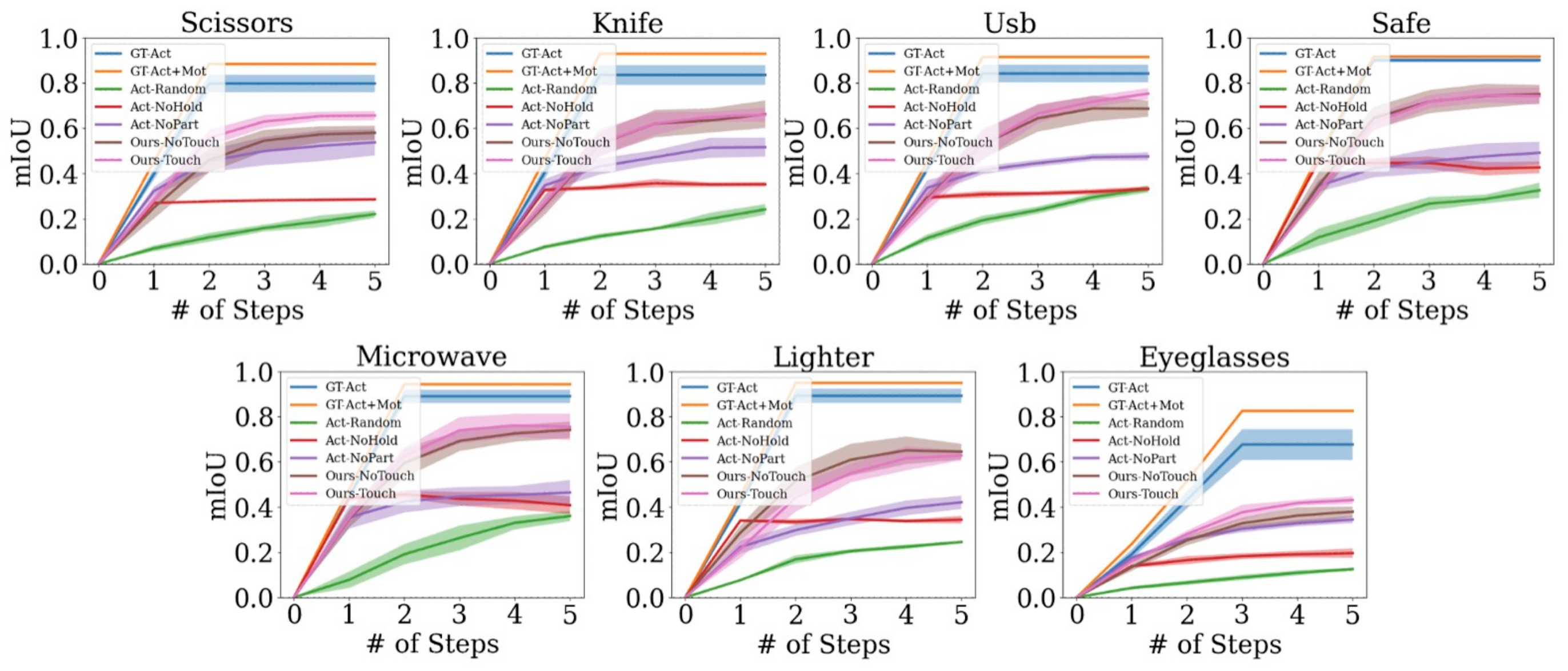}
\caption{\textbf{IoU w.r.t. Interaction Steps.} Results for the remaining simulation categories not shown in the paper due to space restrictions.}
\label{fig:iou_all}
\end{figure*}

\mypara{IoU and Interaction Steps.}
We present simulation benchmark results for the remaining object categories in Fig. \ref{fig:iou_all}.
Our method approaches the upper bounds.

\begin{figure*}[t]
\includegraphics[width=40pc]{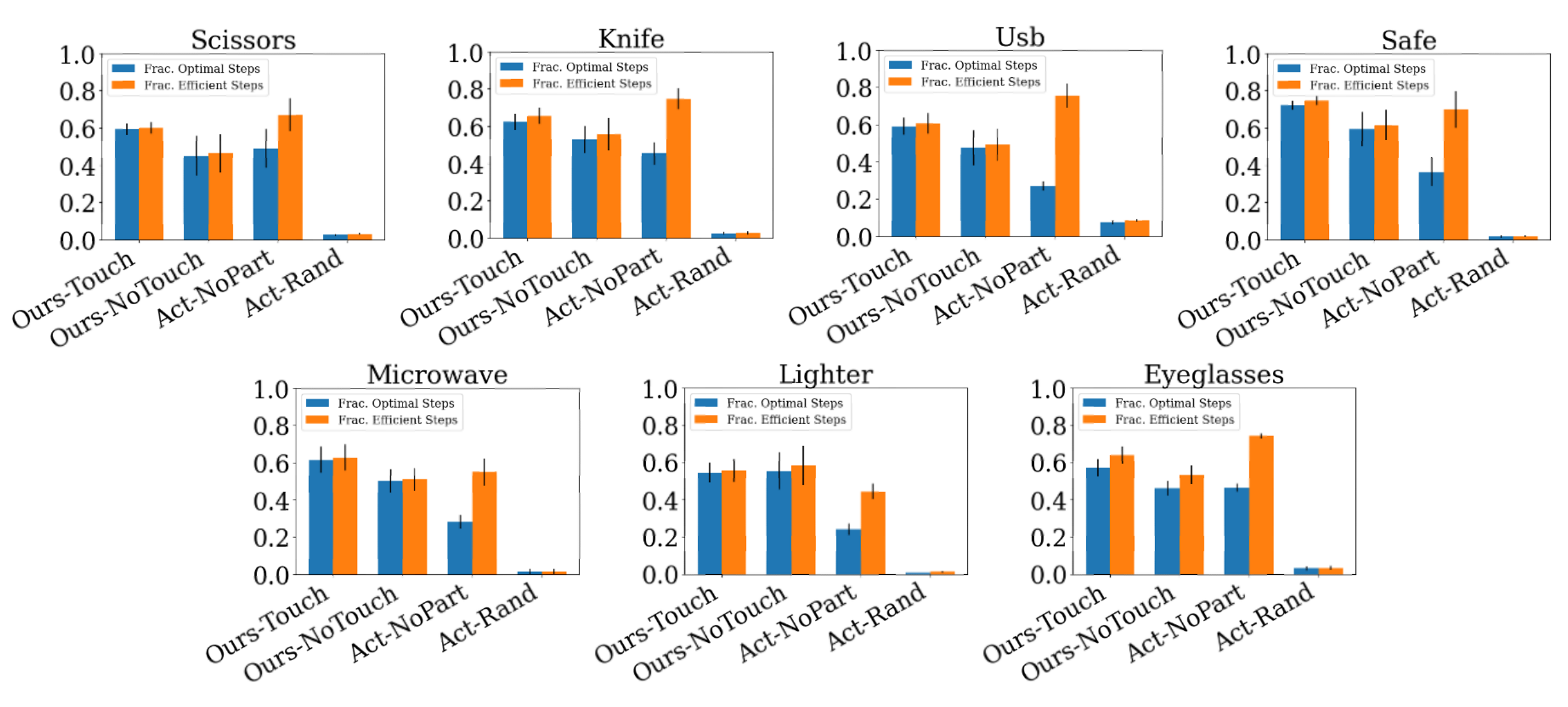}
\caption{\textbf{Effective and Optimal Steps.} Results for the remaining seven simulation categories not shown in the paper due to space restrictions.}
\label{fig:effective_optimal_all}
\end{figure*}

\mypara{Effective and Optimal Actions.}
We notice our [Ours-Touch] and [Act-NoPart] perform efficient actions at roughly the same rate.
However, [Ours-Touch] is able to find may more optimal actions.
See Fig. \ref{fig:effective_optimal_all}.

\end{document}